\documentclass[10pt,twocolumn,letterpaper]{article}

\usepackage{cvpr}              

\usepackage{float}
\usepackage{threeparttable}
\usepackage{algorithm}
\usepackage{algorithmicx}
\usepackage{algpseudocode}
\usepackage[utf8]{inputenc}
\usepackage[T1]{fontenc}
\usepackage{amsfonts}
\usepackage{amsthm}
\usepackage{nicefrac}
\usepackage{tikz}
\usepackage[most]{tcolorbox}
\usetikzlibrary{shapes, arrows.meta, positioning}

\newcounter{prompt}
\newtcolorbox{promptbox}[2][]{%
  enhanced,
  colback=gray!5!white,
  colframe=gray!40!black,
  boxrule=0.5pt,
  arc=2mm,
  left=2mm, right=2mm, top=1mm, bottom=1mm,
  before=\refstepcounter{prompt},
  title={#2},
  #1
}

\definecolor{cvprblue}{rgb}{0.21,0.49,0.74}
\usepackage[pagebackref,breaklinks,colorlinks,allcolors=cvprblue]{hyperref}

\newcommand{\paperID}{EvGenFM-34} 
\def\confName{CVPR-EVGENFM26}
\def\confYear{2026}

\title{DB-3DME: From Dataset to Benchmark for Human-aligned Automatic 3D Mesh Evaluation}

\author{
Nanshan Jia$^{1,2}$ \and
Zhenyu Zhao$^{2}$\thanks{Corresponding authors.} 
\and
Junbo Peng$^{2}$\and
Sui Huang$^{2}$ \and
Jingshen Wang$^{1}$ \and
Zeyu Zheng$^{1}$\footnotemark[1] \\
$^{1}$University of California, Berkeley \\
$^{2}$Roblox Corporation
}

\begin{document}
\maketitle

\begin{abstract}
Recent advances in 3D generation have led to unprecedented improvements in realism, controllability, and efficiency, yet the evaluation of 3D assets remains underexplored. Existing evaluation paradigms, including human evaluation, learned metrics, and vision--language models (VLMs) as judges, suffer from limitations in cost, scalability, resolution handling, or task-specific alignment. In this work, we focus on 3D mesh evaluation and address these challenges by introducing DB-3DME, the Dataset and Benchmark for 3D Mesh Evaluation. We provide a curated dataset of 2,619 synthetic 3D meshes paired with human ratings. Leveraging this dataset, we systematically benchmark the performance of state-of-the-art VLMs and identify visual encoding of 3D representations as a key factor for performance. Building on these insights, we fine-tune an open-source VLM to better align with the requirements of 3D mesh evaluation. Our fine-tuned model substantially outperforms existing pre-trained VLMs across multiple evaluation dimensions. It establishes a new benchmark and provides a foundation for future research on automatic 3D evaluation. We publicly release the benchmark dataset on both GitHub\footnote{\url{https://github.com/nsjia/DB-3DME}} and Hugging Face\footnote{\url{https://huggingface.co/datasets/nsjia/DB-3DME}} to facilitate community usage and future research.
\end{abstract}


\section{Introduction}
\label{sec:intro}

\begin{figure*}[t]
\centering
  \includegraphics[width=0.75\textwidth]{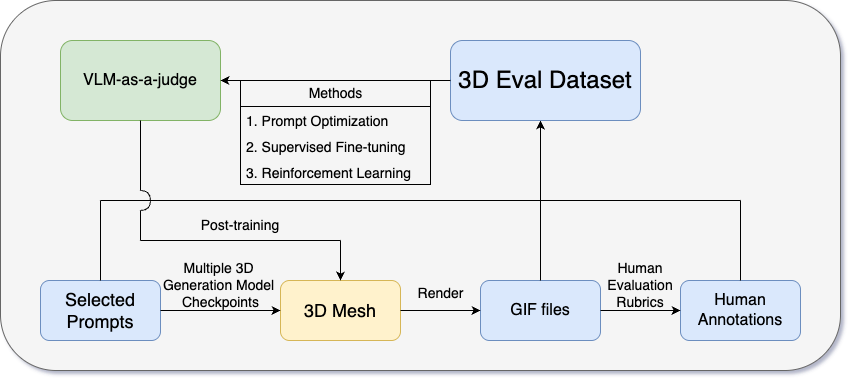}
  \caption{Workflow of 3D mesh evaluation. Blue icons denote the released 3D mesh dataset, the green icon represents the evaluation method, and the yellow icon indicates the processed 3D assets used for evaluation-driven improvement.}
  \label{fig:teaser}
\end{figure*}

With the advent of Cube 3D \cite{roblox2025cube}, 3D object and scene generation has reached an unprecedented level of realism and diversity. A series of other models, such as \cite{jun2023shap,yang2024hunyuan3d,bensadoun2024meta3dgen,hunyuan3d22025tencent,lai2025hunyuan3d25highfidelity3d}, have further revolutionized the 3D generation community, pushing the boundaries of controllability, fidelity, and efficiency to a new stage. While generation methods have rapidly advanced, 3D evaluation remains largely underexplored. In practice, automatic evaluation is crucial not only for assessing the quality of generative models at scale \cite{wu2024gpt}, but also for improving training workflows. For instance, \cite{li2024scalingfilter} leverages automatic evaluation to identify and filter low-quality training samples. Moreover, evaluation models can naturally serve as reward functions for reinforcement learning-based post-training \cite{lee2025calibrated,tamboli2025balanceddpo}. See \cref{fig:teaser} for an overview of the entire evaluation workflow.

Despite these advantages, the current landscape of automatic 3D evaluation remains limited, leaving a substantial gap between the rapid progress of generation methods and the slower development of evaluation techniques. In particular, existing 3D evaluation approaches mainly rely on three paradigms: human evaluation, learned metrics, and VLM-as-a-judge. Human evaluation typically involves expert annotators who assess 3D generations according to a predefined rubric. While this method aligns well with human perception and preferences, it is inherently non-automatic and cannot be easily incorporated into pipelines to distribute. Moreover, human evaluation is costly, as reported by \cite{pan2024human}. This further restricts its applicability in large datasets or iterative model development. Finally, human ratings often suffer from inconsistency, as subjective opinions may introduce bias and variance into the evaluation results.

In contrast, learned metrics and VLM-as-a-judge approaches enable scalable and fully automatic evaluation, provided that inference can be executed efficiently. However, most existing learned metrics rely on CLIP-based encoders, which struggle to process high-resolution inputs. This limitation increasingly hampers their effectiveness, as modern 3D asset generation models are capable of producing highly detailed, high-resolution outputs. Meanwhile, most vision--language models (VLMs) are trained for general multi-modal understanding and reasoning rather than the fine-grained, geometry- and adherence-aware requirements of 3D asset evaluation. Empirically, we find that directly deploying pre-trained VLMs often leads to suboptimal assessment performance. We further analyze this issue in \cref{sec:benchmark}. To mitigate it, a growing body of work~\cite{zhang20253dgen,maiti2025gen3deval} adopts post-training strategies, such as supervised fine-tuning and reinforcement learning with human feedback, to adapt pre-trained VLMs for 3D evaluation tasks. These approaches typically require large-scale, carefully annotated datasets for training and validation prior to deployment.

In this work, we focus on a specific subdomain of 3D evaluation: 3D mesh evaluation. 3D meshes serve as an intermediate representation in most 3D asset generation pipelines, bridging high-level generative models and downstream tasks such as rendering, animation, and physical simulation \cite{roblox2025cube}. In other words, the quality of generated meshes directly affects geometric accuracy, surface consistency, and overall asset usability. This makes robust and scalable mesh-level evaluation a critical component of 3D asset assessment. To address the limitations of existing evaluation approaches, we propose DB-3DME, a dataset and benchmark designed specifically for 3D mesh evaluation. We first construct a curated evaluation dataset consisting of synthetic 3D meshes, each paired with human annotations. The human labels cover two evaluation dimensions: Geometry and Prompt Adherence. These labels enable fine-grained assessment of both structural quality and semantic alignment. The resulting dataset contains 2,619 samples and serves as a foundation for systematic benchmarking as well as post-training of evaluation models. Building on this dataset, we conduct a comprehensive benchmark of multiple state-of-the-art (SOTA) vision--language models (VLMs) on the 3D mesh evaluation task. Our analysis reveals that the capability of the visual encoder in modern VLMs plays a key role in achieving high-quality, human-aligned evaluations. Motivated by this finding, we further fine-tune an open-source VLM to better align with human-centered 3D mesh evaluation objectives. The resulting model achieves substantial performance gains over existing SOTA VLMs and establishes a new benchmark for automatic 3D mesh evaluation.

Our main contributions are summarized as follows:
\begin{enumerate}
\item \textbf{Curated 3D Mesh Evaluation Dataset.}
We release a high-quality dataset containing 3D meshes and corresponding human evaluation ratings, designed to facilitate research on automatic 3D evaluation.
\item \textbf{Benchmark of SOTA VLMs.}
We evaluate a variety of pre-trained VLMs on the proposed dataset and provide a systematic comparison. The analysis demonstrates that the quality of visual encoding for 3D representations is a key factor in the performance of VLM-as-a-judge for 3D evaluation.
\item \textbf{Fine-tuning VLMs for 3D Mesh Evaluation.}
Using Qwen-2.5-VL-7B-Instruct as the base model, we propose a fine-tuning strategy to adapt VLMs for 3D mesh evaluation. Experimental results demonstrate that our fine-tuned model significantly outperforms existing VLMs across all evaluation dimensions, setting a new performance benchmark for future research.
\end{enumerate}

The rest of this paper is organized as follows. \Cref{sec:related} reviews prior work on 3D asset generation and evaluation. \Cref{sec:dataset} presents the step-by-step construction of our labeled 3D mesh dataset. \Cref{sec:benchmark} describes our benchmark setup, compares modern VLMs and a non-VLM baseline on this dataset. \Cref{sec:finetune} introduces our fine-tuning strategy for adapting pre-trained VLMs to 3D mesh evaluation. \Cref{sec:experiments} reports the experimental settings and results, including ablation studies and an analysis of performance stratified by prompt complexity. Finally, \cref{sec:conclusion} concludes the paper.

\section{Related Works}
\label{sec:related}

\subsection{3D Asset Generation}
In the early stages, 3D mesh and asset generation largely relied on Neural Radiance Fields (NeRF) representations \cite{mildenhall2021nerf}. These optimization-based methods gave rise to a series of 3D generation models \cite{poole2022dreamfusion,wang2022clip,shi2023mvdream,tang2023dreamgaussian,metzer2023latent,chen2023fantasia3d,lin2023magic3d,wang2023prolificdreamer,zhang2024text2nerf,lumaGenie2024,csm_cube2025}. However, the quality of the generated meshes was often limited: the extracted meshes tended to be noisy, and the optimization process was typically slow due to its iterative nature. Subsequently, the 3D asset generation community evolved along two main directions. The first stream adopts diffusion models or other continuous-state models as the backbone and extends them to the 3D generation domain \cite{nichol2022point,li2023instant3d,alliegro2023polydiff,lan2024ln3diff,meshy2024,yang2024hunyuan3d,csm_cube2025,hunyuan3d22025tencent,lai2025hunyuan3d25highfidelity3d}. The second stream follows a discrete-token paradigm, treating 3D generation similarly to text generation \cite{chen2024meshxl,siddiqui2024meshgpt,roblox2025cube}.

\subsection{3D Asset Evaluation}
Prior research on 3D asset evaluation can be broadly categorized into three main directions: human evaluation, learned metrics, and VLM-as-a-judge approaches. Human evaluation methods are widely adopted and often rely on large-scale platforms to collect annotations in the form of absolute quality scores or pairwise preference comparisons \cite{ebert20253d,zhang20253dgen}. Despite their reliability, such methods are costly and difficult to scale. Another line of work focuses on learned evaluation metrics, which typically leverage pre-trained encoders such as CLIP \cite{radford2021learning} and train evaluation models using human-annotated data to predict perceptual quality \cite{duggal2025eval3d,zhang20253dgen,zhang2025hi3deval}. In addition to learned evaluators, off-the-shelf pre-trained metrics such as CLIP score are also commonly used as automatic evaluation baselines \cite{he2023t}. While these approaches offer improved scalability compared to human evaluation, their performance is often limited by the representation capacity of the underlying encoders. More recently, VLM-as-a-judge methods have emerged as a promising alternative by directly leveraging pre-trained vision-language models to evaluate 3D assets \cite{wu2024gpt}. Beyond direct prompting, some works further adopt open-source VLMs and improve their alignment with human preferences through post-training or preference optimization. This enables more fine-grained and human-aligned evaluation of 3D generation quality \cite{cai2024gt23d,zhang20253dgen,maiti2025gen3deval}.

\begin{figure*}[t]
    \centering
    \includegraphics[width=0.7\linewidth]{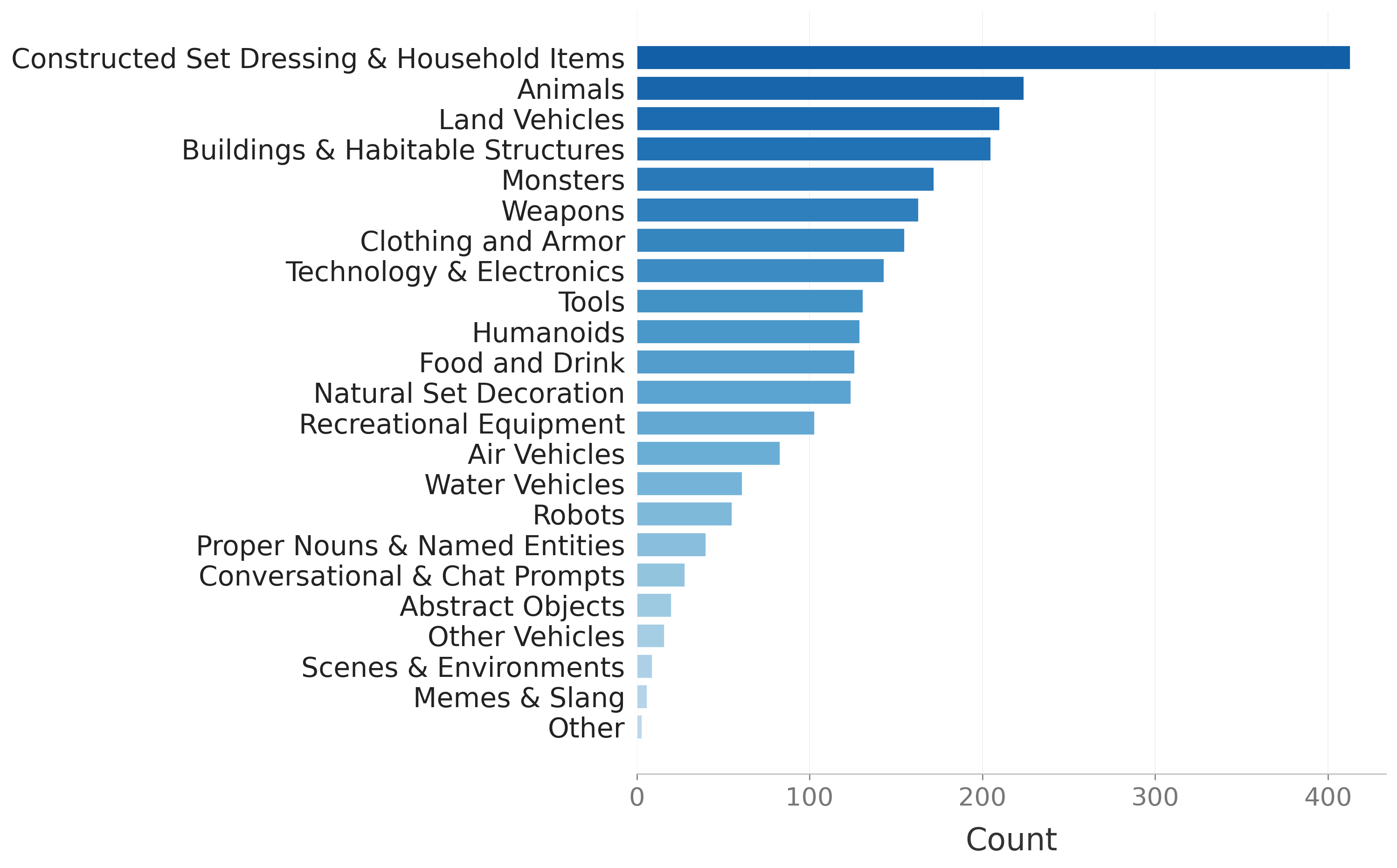}
    \caption{Prompt distribution of our 3D mesh dataset over object categories.}
    \label{fig:prompt_dist}
\end{figure*}

\section{A Curated Dataset for 3D Mesh Evaluation}
\label{sec:dataset}
Our dataset construction follows the workflow in \cref{fig:teaser}. We first curate a set of textual prompts through data analysis on a large-scale online platform. \Cref{sec:prompt_selection} details this prompt sampling process and describes how we generate 3D meshes conditioned on the selected prompts. Next, \cref{sec:dataset_content} outlines our rendering pipeline, including view and storage settings. It also explains the human evaluation procedure used to label meshes on Geometry and Prompt Adherence.

\subsection{Prompt Selection and Mesh Generation}
\label{sec:prompt_selection}
To construct the 3D mesh dataset, we first determine the prompt set. The prompt set is designed to cover a broad range of commonly used prompts while adhering to safety guidelines and avoiding ethical concerns. To ensure both diversity and safety, we analyze user inputs collected from a large-scale online platform. Based on a categorical analysis of this data, we sample 2,619 prompts spanning 22 distinct named categories. \Cref{fig:prompt_dist} shows the distribution of these prompts across the categories. The distribution is intentionally imbalanced to reflect real-world usage patterns observed on the platform. Categories related to everyday objects and environments, such as Constructed Set Dressing \& Household Items, Animals and Land Vehicles, contain the largest number of prompts. This corresponds to their high frequency in user-generated content. In contrast, more specialized categories, including Abstract Objects, Scenes \& Environments, Memes \& Slang, contain fewer prompts. This long-tailed distribution enables coverage of both common and rare semantic concepts. Using the selected prompts, we generate 3D meshes with multiple internal model checkpoints. Additional details on the generation process are provided in \cite{roblox2025cube}.

\subsection{Content of the Dataset}
\label{sec:dataset_content}
In the context of 3D generation, the produced objects or meshes are typically stored in the \texttt{.glb} format. A GLB file is the binary version of the GL Transmission Format (glTF), a modern standard for representing and transmitting 3D assets. It compactly encapsulates all components of a 3D model into a single, efficient binary file. However, current VLMs cannot directly process GLB files or encode them into latent representations for 3D understanding and reasoning. To bridge this gap, each 3D object is instead represented by a set of rendered 2D images captured from multiple viewpoints. Specifically, for each 3D mesh $x$, we render the GLB file into a sequence of $N$ frames forming a GIF animation, denoted as
\begin{equation}
    \mathbf{render}(x) = [x_1, x_2, \ldots, x_N],
\end{equation}
where each $x_i$ represents a 2D screenshot of the 3D mesh from a specific viewing angle. In our dataset, we choose $N=24$ and release only the rendered \texttt{.gif} files for community usage.

To assess the quality of a 3D mesh, we focus on two evaluation dimensions:
\begin{enumerate}
    \item \textbf{Geometry:} measures the structural fidelity and overall shape quality of the 3D mesh.
    \item \textbf{Prompt Adherence:} evaluates how well the 3D mesh aligns with the given textual prompt.
\end{enumerate}

For each sample, the dataset includes the textual prompt and human ratings on both dimensions. Human ratings are restricted to $\{1,2,3\}$. The rubric provided to human annotators is included in the supplementary material.

\begin{figure*}[t]
    \centering
    \begin{minipage}{0.5\linewidth}
        \includegraphics[width=\linewidth]{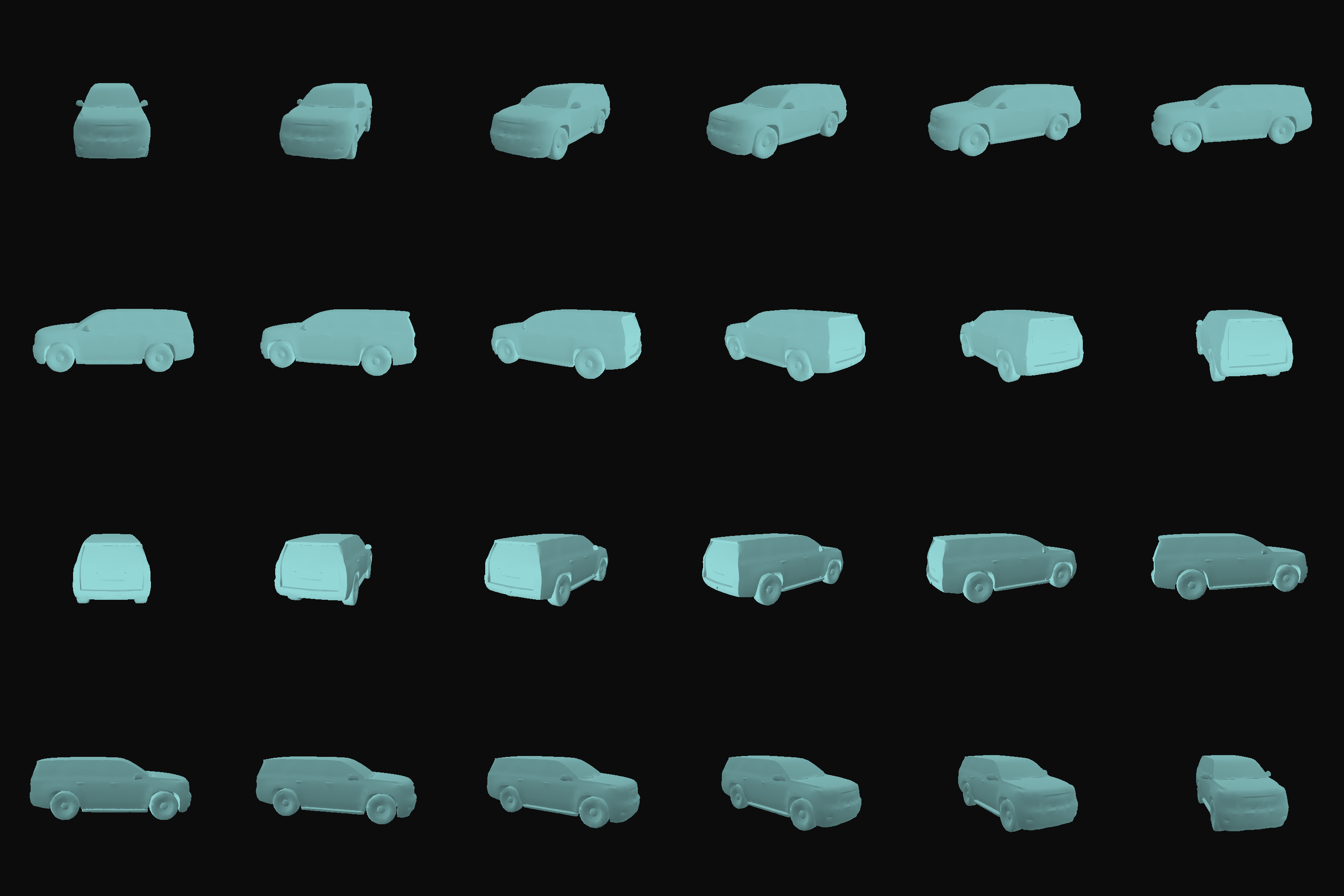}
    \end{minipage}%
    \hfill
    \begin{minipage}{0.38\linewidth}
        \begin{itemize}
            \item \textbf{Prompt:} SUV
            \medskip
            \item \textbf{Geometry:} 2/3
            \medskip
            \item \textbf{Adherence:} 3/3
        \end{itemize}
    \end{minipage}
    \caption{Example sample from our dataset, showing the 3D mesh rendered as a $4 \times 6$ grid of frames (left) and its associated prompt and human ratings (right).}
    \label{fig:suv_example}
\end{figure*}

Finally, we present a sample from the dataset in \cref{fig:suv_example}. We manually collect each frame from the GIF file and organize them into a $4 \times 6$ grid image for presentation. On the right, we show the associated textual prompt, the Geometry rating, and the Prompt Adherence rating. More examples are provided in the supplementary material.

\section{Automatic Evaluation by Prompting Pre-trained Vision Language Models}
\label{sec:benchmark}
In this section, we conduct an automatic evaluation of the released dataset using pre-trained vision-language models (VLMs). Guided by the evaluation rubrics applied to human annotators, we design two versions of system prompts to describe the evaluation task to the VLMs. Both versions share the same task description and evaluation rubrics, instructing the VLM to rate each 3D mesh on Geometry and Prompt Adherence using a scale of $\{1,2,3\}$. The two versions differ in their output format: the first version (V1) requests the VLM to provide ratings directly, while the second version (V2) asks the VLM to provide both ratings and justifications for its ratings. This approach is analogous to the Chain-of-Thought (CoT) technique widely used in reasoning with large language models \cite{wei2022chain}. The full version of the system prompts are provided in the supplementary material. During evaluation, the GIF file for 3D mesh is first converted to multiple images and then transformed to a large grid image. The grid image is sent to the VLM together with the system prompt and the textual prompt of the mesh. For more technical details, please refer to the supplementary material.

For the VLMs, we select the following baseline models:
\begin{itemize}
    \item \textbf{ChatGPT-4o.}
    \item \textbf{ChatGPT-4.1.}
    \item \textbf{ChatGPT-5 series.} We evaluate three variants corresponding to different levels of reasoning effort, including low level, medium level and high level. These variants trade off inference cost for reasoning depth and are denoted as \emph{ChatGPT-5-low}, \emph{ChatGPT-5-medium}, and \emph{ChatGPT-5-high}.
    \item \textbf{Gemini-2.5-Pro.}
    \item \textbf{Qwen-2.5-VL series.} We primarily evaluate the 7B variant.
    \item \textbf{CLIP Score~\cite{radford2021learning}.} As a non-VLM baseline, we compute the CLIP cosine similarity between each rendered grid image and its textual prompt using ViT-L/14. Since CLIP Score produces a continuous value rather than a discrete rating, we report only Pearson correlation.
\end{itemize}

To compare the quality of these automatic evaluation models, we consider three metrics: \emph{Exact-Match Rate}, \emph{Within-One Rate} and \emph{Pearson correlation}. The Exact-Match Rate represents the proportion of samples for which the VLM ratings perfectly match human ratings, while the Within-One Rate indicates the proportion of samples where the VLM ratings deviate from human ratings by at most one point. The Pearson correlation measures the linear correlation coefficient between the VLM ratings and the corresponding human labels. Performance of the VLM-as-a-judge on Geometry and Prompt Adherence are reported in \cref{tab:geometry_vlm_comparison} and \cref{tab:prompt_adherence_vlm_comparison}, respectively.

\begin{table}[t]
\centering
\caption{Comparison of different VLMs on the evaluation of Geometry.
``Exact-Match'' indicates the proportion of samples for which the VLM ratings perfectly match human labels, while ``Within-One'' measures the proportion of samples where the VLM ratings deviate from human labels by no more than one point. ``Pearson'' measures the linear correlation between VLM ratings and human labels. V1 prompts the VLM to directly output ratings, whereas V2 additionally elicits justifications \cite{wei2022chain}.}
\label{tab:geometry_vlm_comparison}
\resizebox{\linewidth}{!}{%
\begin{tabular}{llccc}
\toprule
Version & VLM & Exact-Match & Within-One & Pearson\\
\midrule
\multicolumn{2}{l}{\emph{Non-VLM baseline}} \\
-- & CLIP Score (ViT-L/14) & -- & -- & 0.074\\
\midrule
V1 & ChatGPT-4o & 0.377 & 0.854 & 0.260\\
 & Qwen-2.5-vl-7b & 0.439 & 0.878 & 0.176\\
 & ChatGPT-5-high & 0.353 & 0.827 & 0.269 \\
 & ChatGPT-5-medium & 0.361 & 0.826 & 0.260\\
 & ChatGPT-5-low & 0.358 & 0.828 & 0.256\\
 & ChatGPT-4.1 & 0.385 & 0.861 & 0.226\\
 & Gemini-2.5-pro & \textbf{0.445} & 0.891 & \textbf{0.370}\\
\midrule
V2 & ChatGPT-4o & 0.384 & 0.864 & 0.269\\
 & Qwen-2.5-vl-7b & 0.424 & 0.868 & 0.174\\
 & ChatGPT-5-high & 0.356 & 0.832 & 0.258\\
 & ChatGPT-5-medium & 0.369 & 0.841 & 0.279\\
 & ChatGPT-5-low & 0.361 & 0.839 & 0.269\\
 & ChatGPT-4.1 & 0.389 & 0.867 & 0.223\\
 & Gemini-2.5-pro & 0.438 & \textbf{0.897} & 0.369\\
\bottomrule
\end{tabular}%
}
\end{table}

\begin{table}[t]
\centering
\caption{Comparison of different VLMs on the evaluation of Prompt Adherence.
``Exact-Match'' indicates the proportion of samples for which the VLM ratings perfectly match human labels, while ``Within-One'' measures the proportion where VLM ratings deviate by no more than one point. ``Pearson'' measures the linear correlation. V1 prompts the VLM to directly output ratings, whereas V2 additionally elicits justifications \cite{wei2022chain}.}
\label{tab:prompt_adherence_vlm_comparison}
\resizebox{\linewidth}{!}{%
\begin{tabular}{llccc}
\toprule
Version & VLM & Exact-Match & Within-One & Pearson\\
\midrule
\multicolumn{2}{l}{\emph{Non-VLM baseline}} \\
-- & CLIP Score (ViT-L/14) & -- & -- & 0.228\\
\midrule
V1 & ChatGPT-4o & 0.644 & 0.940 & 0.414\\
 & Qwen-2.5-vl-7b & 0.500 & 0.872 & 0.232\\
 & ChatGPT-5-high & \textbf{0.649} & 0.956 & 0.459\\
 & ChatGPT-5-medium & 0.639 & 0.956 & 0.451\\
 & ChatGPT-5-low & 0.640 & 0.945 & 0.427\\
 & ChatGPT-4.1 & 0.605 & 0.939 & 0.393\\
 & Gemini-2.5-pro & 0.648 & 0.940 & 0.454\\
\midrule
V2 & ChatGPT-4o & 0.641 & 0.942 & 0.413\\
 & Qwen-2.5-vl-7b & 0.507 & 0.837 & 0.214\\
 & ChatGPT-5-high & 0.643 & \textbf{0.964} & \textbf{0.483}\\
 & ChatGPT-5-medium & 0.637 & 0.963 & 0.468\\
 & ChatGPT-5-low & 0.645 & 0.954 & 0.472\\
 & ChatGPT-4.1 & 0.602 & 0.944 & 0.404\\
 & Gemini-2.5-pro & 0.636 & 0.940 & 0.452\\
\bottomrule
\end{tabular}%
}
\end{table}

From these results, we make the following observations. First, system prompt V2 does not provide a significant performance improvement over system prompt V1. A similar finding, where Chain-of-Thought (CoT) prompting fails to enhance performance on non-mathematical or non-symbolic reasoning tasks, has also been reported in \cite{sprague2024cot}. Second, within the ChatGPT-5 series, variants with higher reasoning effort do not yield notable improvements in either Geometry or Prompt Adherence compared to those with lower reasoning effort. Taken together, these two points suggest that simply increasing reasoning effort does not lead to better alignment, and reasoning capability is not the primary bottleneck for aligning automatic evaluation with human ratings.

Furthermore, VLMs consistently exhibit stronger alignment with human annotators on Prompt Adherence than on Geometry. Since Geometry is an evaluation dimension grounded purely in the visual modality, whereas Prompt Adherence involves both visual and textual modalities, this performance gap emphasizes the importance of visual processing. In summary, we conclude that \textbf{effectiveness of visual encoding is an important factor that influences VLM-as-a-judge performance in 3D evaluation}.

\paragraph{Comparison with non-VLM metrics.}
We include CLIP Score~\cite{radford2021learning} as a non-VLM baseline that measures cosine similarity between the rendered image and the text prompt. Because CLIP Score produces a continuous value, we compare using Pearson correlation rather than Exact-Match or Within-One. As shown in \cref{tab:geometry_vlm_comparison,tab:prompt_adherence_vlm_comparison}, CLIP Score achieves a Pearson correlation of only $0.074$ on Geometry and $0.228$ on Prompt Adherence, substantially below all VLM baselines. This gap highlights the limitations of generic image--text similarity for the nuanced, rubric-based assessment captured in our benchmark. We note that other 3D-specific metrics (\eg, Eval3D, Hi3DEval, 3DGen-Bench) similarly produce continuous scores and are designed for model-level rankings rather than per-sample discrete ratings; a direct comparison in the $\{1,2,3\}$ framework would require arbitrary binning, so we limit the non-VLM comparison to the correlation-based CLIP baseline.

\section{Adapting Open-source Vision-Language Models via Fine-tuning}
\label{sec:finetune}
To improve the visual processing capability identified in \cref{sec:benchmark}, we focus on improving the visual encoder in open-source VLMs. Specifically, we fine-tune the visual encoder while freezing the language model component. To enable gradient backpropagation, the entire workflow must remain differentiable. Instead of using the language model head to obtain decoding logits, we extract the last hidden state corresponding to the final token. This hidden state is believed to capture the semantic context of the model's internal ``working memory'' immediately before prediction. We then introduce a score head that maps this final hidden state to a scalar output, which is interpreted as the predicted rating. To align the predicted scores with human ratings, we explore two loss functions:

\begin{enumerate}
    \item \textbf{Mean Squared Error (MSE).}
    The MSE loss measures the squared $\ell_2$ distance between the predicted score $\hat{y}_{\mathrm{head}}$ and the corresponding human rating $y$, defined as
    \begin{equation}
        \mathcal{L}_{\mathrm{MSE}} = \frac{1}{B}\sum_{i=1}^{B}(\hat{y}_{\mathrm{head},i} - y_i)^2,
    \end{equation}
    where $B$ denotes the batch size, and $i$ indexes the samples within the batch.

    \item \textbf{Cross-Entropy (CE).}
    To define the CE loss, we first convert the scalar output $\hat{y}_{\mathrm{head}}$ into logits. Specifically,
    \begin{align*}
        \mathrm{logits} =& [\mathrm{logits}_1, \mathrm{logits}_2, \mathrm{logits}_3] \\
        = &
        [-\alpha(\hat{y}_{\mathrm{head}} - 1)^2, -\alpha(\hat{y}_{\mathrm{head}} - 2)^2, -\alpha(\hat{y}_{\mathrm{head}} - 3)^2],
    \end{align*}
    where $\alpha > 0$ is a hyperparameter controlling the sharpness of the logits. In essence, the logits are defined by the negative squared distance between the predicted score and each discrete rating level.
    These logits are then transformed into probabilities using the $\mathrm{Softmax}$ function:
    \begin{equation}
        p = \mathrm{Softmax}(\mathrm{logits}) = [p_1, p_2, p_3],\label{eq:logits-prob}
    \end{equation}
    where \[ \quad
        p_i = \frac{e^{\mathrm{logits}_i}}{\sum_{j=1}^{3} e^{\mathrm{logits}_j}}, \quad 1\leq i \leq 3.\]
    The final CE loss is given by
    \begin{equation}
        \mathcal{L}_{\mathrm{CE}} = -\frac{1}{B}\sum_{i=1}^{B}\sum_{j=1}^{3}\mathbf{1}(y_i = j)\log p_{j,i},
    \end{equation}
    where $B$ is the batch size, $y_i$ denotes the human rating for the $i$-th sample, $p_{j,i}$ represents the predicted probability of class $j$ following \cref{eq:logits-prob} for the $i$-th sample, and $\mathbf{1}(\cdot)$ is the indicator function.
\end{enumerate}

During inference, once we obtain the output scalar $\hat{y}_{\mathrm{head}}$, we discretize it by rounding to the nearest discrete rating in the rating set $\{1,2,3\}$, \ie,
\begin{equation}
    \hat{y}_{\mathrm{vlm}} =
    \begin{cases}
        1, & \text{if } \hat{y}_{\mathrm{head}} < 1.5, \\
        2, & \text{if } 1.5 \leq \hat{y}_{\mathrm{head}} < 2.5, \\
        3, & \text{if } \hat{y}_{\mathrm{head}} \geq 2.5.
    \end{cases}
\end{equation}

\section{Experiments}
\label{sec:experiments}

\subsection{Experimental Settings}
In our experiments, we use our released dataset and randomly split it into training and test sets with an 8:2 ratio, using a fixed random seed of 42. The same split is consistently applied across all experiments to ensure comparability. For the pre-trained VLM, we adopt Qwen-2.5-VL-7B \cite{bai2025qwen25vltechnicalreport}. We apply LoRA \cite{hu2022lora} to fine-tune only the visual encoder and keep the language model frozen. On top of the model, we use a linear layer as the score head that maps the last hidden state of the final token to a scalar prediction. The model is fine-tuned separately on Geometry and on Prompt Adherence.

Modern vision-language models generally process images by patchifying them into fixed-size tokens using a ViT-like encoder \cite{dosovitskiy2020image}. Consequently, the sequence length for the visual input scales approximately linearly with the total number of image patches, or equivalently, quadratically with the input resolution. To ensure efficient fine-tuning and control memory usage, we limit each sample to 12 frames. We select every other frame from the first 24 frames (\ie, frames 1, 3, \ldots, 23), resize each frame from $512 \times 512$ to $256 \times 256$, and arrange them into a $3 \times 4$ grid. This results in a single grid image of size $768 \times 1024$ that is used as input to the model. For Geometry, the input consists of the grid image of the sample paired with the text ``Rate the Geometry from 1 to 3''. For Prompt Adherence, the input includes both the grid image and its corresponding prompt, followed by ``Rate the Prompt Adherence from 1 to 3''. Note that unlike the setup in \cref{sec:benchmark}, we do not use system prompts during training or inference. This design has two benefits. First, it reduces the length of the textual sequence, so that the overall efficiency of training and inference are improved. Second, it replaces the subjective task descriptions and rubrics by an objective sentence. This helps alleviate potential subjective bias from human evaluations.

\subsection{Benchmark Comparison}
We compare the performance of the fine-tuned models with other benchmark models in this section. Since the training set has been used for fine-tuning, all evaluations are conducted exclusively on the test set. For the benchmark models, we report only the better result between system prompt V1 and system prompt V2. Consistent with \cref{sec:benchmark}, we use Exact-Match rate, Within-One rate and Pearson correlation to assess model performance. We additionally present the half-width of the $95\%$ confidence interval of the Exact-Match rate. The experimental results are presented in \cref{tab:geometry_results} and \cref{tab:prompt_adherence_results}.

\begin{table}[t]
\centering
\caption{Comparison of VLMs on Geometry evaluation. ``Exact Match'' indicates the proportion of samples where VLM ratings perfectly match human labels. ``Within One'' measures the proportion where VLM ratings deviate by no more than one point. ``Pearson'' measures the linear correlation. We report the better result between the two system prompt versions for all pre-trained models.}
\resizebox{\linewidth}{!}{%
\begin{tabular}{lccc}
\toprule
VLM & Exact-Match & Within-One & Pearson\\
\midrule
ChatGPT-4o & $0.365 \pm 0.041$ & 0.870 & 0.282\\
Qwen-2.5-VL-7B & $0.439 \pm 0.042$ & 0.893 & 0.253\\
ChatGPT-5-high & $0.340 \pm 0.041$ & 0.840 & 0.274\\
ChatGPT-5-medium & $0.352 \pm 0.041$ & 0.847 & 0.271\\
ChatGPT-5-low & $0.342 \pm 0.041$ & 0.840 & 0.281\\
ChatGPT-4.1 & $0.403 \pm 0.042$ & 0.882 & 0.289\\
Gemini-2.5-pro & $0.431 \pm 0.042$ & 0.895 & 0.346\\
Ours (fine-tuned) & $\mathbf{0.618} \pm 0.042$ & \textbf{0.987} & \textbf{0.594}\\
\bottomrule
\end{tabular}%
}
\label{tab:geometry_results}
\end{table}

\begin{table}[t]
\centering
\caption{Comparison of VLMs on Prompt Adherence evaluation. ``Exact Match'' indicates the proportion of samples where VLM ratings perfectly match human labels. ``Within One'' measures the proportion where VLM ratings deviate by no more than one point. ``Pearson'' measures the linear correlation. We report the better result between the two system prompt versions for all pre-trained models.}
\resizebox{\linewidth}{!}{%
\begin{tabular}{lccc}
\toprule
VLM & Exact-Match & Within-One & Pearson\\
\midrule
ChatGPT-4o & $0.647 \pm 0.041$ & 0.954 & 0.463\\
Qwen-2.5-VL-7B & $0.504 \pm 0.043$ & 0.868 & 0.207\\
ChatGPT-5-high & $0.679 \pm 0.040$ & \textbf{0.969} & 0.542\\
ChatGPT-5-medium & $0.654 \pm 0.041$ & 0.964 & 0.499\\
ChatGPT-5-low & $0.662 \pm 0.040$ & 0.950 & 0.489\\
ChatGPT-4.1 & $0.608 \pm 0.042$ & 0.956 & 0.445 \\
Gemini-2.5-pro & $0.651 \pm 0.041$ & 0.952 & 0.490\\
Ours (fine-tuned) & $\mathbf{0.695} \pm 0.039$ & \textbf{0.969} & \textbf{0.583}\\
\bottomrule
\end{tabular}%
}
\label{tab:prompt_adherence_results}
\end{table}

From the results, we observe that the fine-tuned Qwen-2.5-VL-7B model substantially outperforms the base model and other benchmark models on Geometry, while it slightly exceeds the performance of benchmark models on Prompt Adherence. This pattern suggests that, with proper fine-tuning, the visual processing can be enhanced to adapt to the 3D evaluation tasks. Additionally, the limited gain of the fine-tuned model on Prompt Adherence compared to top-performing models can be attributed to model size constraints. While ChatGPT-5 and other large models benefit from their higher capacity, our model has only 7B parameters, yet it achieves competitive performance with potentially better efficiency. Overall, the fine-tuned model establishes a new reference point for both Geometry and Prompt Adherence, serving as a benchmark for the research community.

\subsection{Ablations}
\label{sec:ablations}
In this section, we conduct ablation studies to validate our design choice of fine-tuning the vision encoder while freezing the language model. Using Geometry as a representative task, we evaluate several training configurations that differ in which components of the VLM are fine-tuned, as summarized in \cref{tab:ablation}. Specifically, we consider:
\begin{enumerate}
    \item \textbf{Original model (no training).} The pre-trained Qwen-2.5-VL-7B is used directly for inference, without any fine-tuning or additional training, and ratings are extracted from the textual output.
    \item \textbf{Score head only.} The entire VLM is frozen, and a score head is trained to replace the LM head for rating prediction.
    \item \textbf{Vision encoder + score head.} The vision encoder is fine-tuned with LoRA while jointly training the score head.
    \item \textbf{Language model + score head.} The language model is fine-tuned with LoRA while jointly training the score head.
    \item \textbf{Vision encoder + language model + score head.} Both the vision encoder and language model are fine-tuned with LoRA while jointly training the score head.
\end{enumerate}

\begin{table}[t]
\centering
\caption{Ablation study of trainable modules for Geometry evaluation. The ``original model'' performs inference with pre-trained Qwen-2.5-VL-7B. The ``score head'' setting freezes the VLM and replaces the LM head with a score head. Other settings fine-tune the vision encoder and/or language model with LoRA while jointly training the score head.}
\resizebox{\linewidth}{!}{%
\begin{tabular}{lccc}
\toprule
Fine-tuned module & Exact-Match & Within-One & Pearson\\
\midrule
Original model (no fine-tuning) & 0.439 & 0.891 & 0.230\\
Score head & 0.460 & \textbf{0.996} & 0.011\\
Vision enc.\ + score head & \textbf{0.618} & 0.987 & \textbf{0.594}\\
Language model + score head & 0.567 & 0.990 & 0.519\\
Vision enc.\ + LM + score head & 0.605 & 0.977 & 0.554\\
\bottomrule
\end{tabular}%
}
\label{tab:ablation}
\end{table}

From these results, we make the following observations. First, the pre-trained model and the setting that trains only the score head achieve similar exact-match rates. This indicates that \textbf{the performance gains of fine-tuned models come from the updated model weights learned during training}, rather than from replacing the original decoding strategy with a score head.

Second, fine-tuning the vision encoder together with the score head leads to the best performance in terms of exact-match rate and Pearson correlation. This result highlights \textbf{the importance of adapting visual representations to specific evaluation tasks}.

Third, further fine-tuning the language model on top of the vision encoder does not yield additional improvements in any metric. Despite having more trainable parameters, the performance remains comparable. This suggests limited benefit from adapting the language model once the vision encoder is fine-tuned.

Fourth, fine-tuning the language model with the score head, while keeping the vision encoder frozen, performs worse than fine-tuning both the vision encoder and language model on exact-match rate and Pearson correlation. This result indicates that \textbf{adapting the language model alone cannot compensate for a frozen vision encoder}.

Overall, these ablation findings suggest that strengthening the performance of the vision encoder is vital to the 3D evaluation performance in current settings. They also justify our design choice to fine-tune the vision encoder while freezing the language model in \cref{sec:finetune}.

\subsection{Analysis by Prompt Complexity}
\label{sec:prompt_complexity}
To provide finer-grained insight into VLM evaluation, we partition the test set by prompt complexity: \emph{Short} prompts ($\le5$ words, $n\!=\!393$) and \emph{Detailed} prompts ($\ge6$ words, $n\!=\!131$). \Cref{tab:prompt_complexity} shows the performance of representative VLMs on Prompt Adherence stratified by prompt length.

\begin{table}[t]
\centering
\caption{Prompt Adherence performance stratified by prompt complexity. ``Short'' denotes prompts with $\le5$ words and ``Detailed'' denotes $\ge6$ words. We report the best-performing prompt version for each pre-trained model.}
\label{tab:prompt_complexity}
\resizebox{\linewidth}{!}{%
\begin{tabular}{l cc cc}
\toprule
 & \multicolumn{2}{c}{Exact-Match} & \multicolumn{2}{c}{Pearson} \\
\cmidrule(lr){2-3} \cmidrule(lr){4-5}
VLM & Short & Detailed & Short & Detailed \\
\midrule
ChatGPT-4o      & 0.672 & 0.573 & 0.441 & 0.504 \\
ChatGPT-5-high  & 0.700 & 0.618 & 0.538 & 0.543 \\
ChatGPT-5-medium & 0.689 & 0.550 & 0.530 & 0.397 \\
Gemini-2.5-pro  & 0.667 & 0.603 & 0.474 & 0.519 \\
\midrule
Ours (fine-tuned) & \textbf{0.718} & \textbf{0.626} & \textbf{0.599} & \textbf{0.550} \\
\bottomrule
\end{tabular}%
}
\end{table}

All models achieve higher Exact-Match rates on short prompts, which is expected given the simpler semantic content. Interestingly, for Pearson correlation, this pattern reverses in some cases: ChatGPT-4o and ChatGPT-5-high achieve higher correlation on detailed prompts. This suggests that while absolute agreement is easier for short prompts, VLMs may capture the relative ordering of quality more faithfully when the prompt specifies richer visual attributes for assessment.

Notably, our fine-tuned model achieves the highest Exact-Match and Pearson correlation on both short and detailed prompts, outperforming all pre-trained VLMs in every cell of \cref{tab:prompt_complexity}. The fine-tuned model follows the same trend of higher Exact-Match on short prompts ($0.718$ vs.\ $0.626$), while maintaining strong Pearson correlation across both subsets ($0.599$ and $0.550$). This consistent advantage across prompt complexities demonstrates that the fine-tuning strategy generalizes well beyond simple prompts and is not merely memorizing easy cases.

\subsection{Role of Instruction and Language Model in Geometry Scoring}
Building upon the ablation results in \cref{sec:ablations}, we further investigate whether the instruction text and the language model forward pass are fundamentally necessary for visual geometry scoring.

In the fine-tuning setup of \cref{sec:ablations}, the model receives the grid image together with the fixed instruction \texttt{Rate the Geometry from 1 to 3}. This instruction is identical for every sample, so it carries no sample-specific information; the scoring rubric that zero-shot prompting must spell out in a system prompt could instead be learned implicitly from the rating labels. We therefore hypothesize that \textbf{the instruction is unnecessary for Geometry scoring}.

Moreover, the Geometry criterion is a property of the rendered mesh alone: unlike Prompt Adherence, whose rating depends on the per-sample text prompt, a Geometry rating is fully determined by the input image. The vision encoder already encodes this image, whereas the language model in a VLM mainly serves to fuse text with vision and to generate---neither of which is required here. We therefore further hypothesize that \textbf{the language-model forward pass is also unnecessary, and that the vision-encoder features alone suffice}. We test both hypotheses below.

We compare three settings that feed an identical linear score head, ordered by increasing machinery (\cref{tab:readouts_e}):
\begin{itemize}
\item \textbf{A (vision only).} The mean-pooled output of the vision encoder + merger. The language model is bypassed entirely, and no text enters this path.
\item \textbf{B (\,= A's tokens through the LM).} The visual tokens are processed by the frozen language model, and the head reads the last-layer hidden state at the final position --- a last-token readout that has access to the entire image but not the instruction.
\item \textbf{B$^{+}$ (\,= B + instruction).} The instruction (``Rate the Geometry from 1 to 3'') is appended and the head reads the hidden state at its final token, which is the score-head setting in Section~6.3.
\end{itemize}
The step A$\,\rightarrow\,$B isolates the net effect of language-model processing, and B$\,\rightarrow\,$B$^{+}$ the net contribution of instruction conditioning. Each head is a single linear layer trained with mean squared error (AdamW, learning rate $10^{-3}$, weight decay $10^{-4}$, $1{,}000$ full-batch epochs, cosine schedule). To obtain reliable estimates we use 200-trial Monte-Carlo cross-validation: for each trial we draw an independent 80/20 split, with all three settings sharing the split so that comparisons are paired. Because the trials share overlapping training data, we report paired $t$-tests with the Nadeau--Bengio correction.

\begin{table}[t]
\centering
\caption{Results of the three settings on Geometry (Qwen2.5-VL-7B).}
\label{tab:readouts_e}
\resizebox{\linewidth}{!}{%
\begin{tabular}{lcccc}
\toprule
 & LM & Instr. & Exact-Match & Pearson \\
\midrule
A & --- & --- & $0.557 \pm 0.020$ & $0.471 \pm 0.029$ \\
B & \checkmark & --- & $0.492 \pm 0.022$ & $0.337 \pm 0.039$ \\
B$^{+}$ & \checkmark & \checkmark & $0.499 \pm 0.021$ & $0.345 \pm 0.037$ \\
\bottomrule
\end{tabular}}
\end{table}

\begin{table}[t]
\centering
\caption{Effect of adding each component (Qwen2.5-VL-7B).}
\label{tab:chain_e}
\resizebox{\linewidth}{!}{%
\begin{tabular}{lcccc}
\toprule
 & \multicolumn{2}{c}{Exact-Match} & \multicolumn{2}{c}{Pearson} \\
\cmidrule(lr){2-3} \cmidrule(lr){4-5}
 & $\Delta$ & $p$ & $\Delta$ & $p$ \\
\midrule
A$\,\rightarrow\,$B \; (add LM)           & $-0.066$ & $3{\times}10^{-7}$ & $-0.134$ & $1{\times}10^{-10}$ \\
B$\,\rightarrow\,$B$^{+}$ \; (add instr.) & $+0.008$ & $0.59$ & $+0.009$ & $0.73$ \\
\midrule
A$\,\rightarrow\,$B$^{+}$ & $-0.058$ & $4{\times}10^{-6}$ & $-0.126$ & $1{\times}10^{-8}$ \\
\bottomrule
\end{tabular}}
\end{table}

\begin{table}[t]
\centering
\caption{Results of \textbf{Exact-Match} across four Qwen-VL models.}
\label{tab:fivemodels_em_e}
\resizebox{\linewidth}{!}{%
\begin{tabular}{lccc cc cc}
\toprule
 & & & & \multicolumn{2}{c}{A$\rightarrow$B (add LM)} & \multicolumn{2}{c}{B$\rightarrow$B$^{+}$ (add instr.)} \\
\cmidrule(lr){5-6} \cmidrule(lr){7-8}
Model & A & B & B$^{+}$ & $\Delta$ & $p$ & $\Delta$ & $p$ \\
\midrule
Qwen2-VL-2B    & 0.545 & 0.512 & 0.502 & $-0.033$ & $0.0025$            & $-0.010$ & $0.39$ \\
Qwen2.5-VL-3B  & 0.537 & 0.500 & 0.490 & $-0.037$ & $8{\times}10^{-4}$  & $-0.010$ & $0.32$ \\
Qwen2.5-VL-7B  & 0.557 & 0.492 & 0.499 & $-0.066$ & $3{\times}10^{-7}$  & $+0.008$ & $0.59$ \\
Qwen3-VL-4B    & 0.563 & 0.460 & 0.483 & $-0.103$ & $8{\times}10^{-13}$ & $+0.023$ & $0.12$ \\
\bottomrule
\end{tabular}}
\end{table}

\begin{table}[t]
\centering
\caption{Results of \textbf{Pearson} across four Qwen-VL models.}
\label{tab:fivemodels_pr_e}
\resizebox{\linewidth}{!}{%
\begin{tabular}{lccc cc cc}
\toprule
 & & & & \multicolumn{2}{c}{A$\rightarrow$B (add LM)} & \multicolumn{2}{c}{B$\rightarrow$B$^{+}$ (add instr.)} \\
\cmidrule(lr){5-6} \cmidrule(lr){7-8}
Model & A & B & B$^{+}$ & $\Delta$ & $p$ & $\Delta$ & $p$ \\
\midrule
Qwen2-VL-2B    & 0.448 & 0.376 & 0.344 & $-0.072$ & $1{\times}10^{-4}$  & $-0.032$ & $0.13$ \\
Qwen2.5-VL-3B  & 0.429 & 0.325 & 0.311 & $-0.104$ & $2{\times}10^{-6}$  & $-0.014$ & $0.49$ \\
Qwen2.5-VL-7B  & 0.471 & 0.337 & 0.345 & $-0.134$ & $1{\times}10^{-10}$ & $+0.009$ & $0.73$ \\
Qwen3-VL-4B    & 0.483 & 0.293 & 0.300 & $-0.190$ & $1{\times}10^{-13}$ & $+0.007$ & $0.80$ \\
\bottomrule
\end{tabular}}
\end{table}

Appending the task instruction and reading its final token (B$\,\rightarrow\,$B$^{+}$) leaves both metrics essentially unchanged on Qwen2.5-VL-7B (Pearson $+0.009$, $p{=}0.73$; Exact-Match $+0.008$, $p{=}0.59$; \cref{tab:chain_e}), and the effect is never significant across the four models (all $p \geq 0.12$, with inconsistent sign; \cref{tab:fivemodels_em_e,tab:fivemodels_pr_e}). \textbf{The task instruction can thus be dropped without loss,} as anticipated: since the instruction is identical across samples and the rubric is learned from the labels, the task-conditioned ``working memory'' that motivates reading the final token offers nothing beyond an instruction-blind summary of the image.

Removing the language model, by contrast, \emph{raises} the score: the vision-only setting A outperforms the LM readout B on both metrics in every model (all $p < 0.003$; \cref{tab:chain_e,tab:fivemodels_em_e,tab:fivemodels_pr_e}). \textbf{A linear head on the vision encoder + merger is the strongest frozen variant, and skips the language-model forward pass entirely.} That the visual features alone already suffice matches our hypothesis that the vision encoder carries the information critical for Geometry scoring. The drop caused by \emph{adding} the language model is surprising; we speculate it arises because the frozen language model is optimized for next-token prediction rather than visual scoring, so its layers progressively reorient the visual tokens toward the output vocabulary and away from the linearly decodable geometric detail that the score head relies on. This reasoning is specific to image-only criteria; for Prompt Adherence the per-sample text prompt is informative, so a text--vision fusion stage remains necessary, which we leave to future work.

\section{Conclusion}
\label{sec:conclusion}
In this paper, we introduce a curated dataset tailored for 3D mesh evaluation. Building upon the dataset, we benchmark the performance of pre-trained VLMs on the 3D mesh evaluation task and conclude that the performance of vision encoders is critical for high-quality automatic evaluation with VLMs. Inspired by this finding, we further fine-tune the vision encoder of the Qwen2.5-VL-7B model. The fine-tuned model serves as a new benchmark for the 3D mesh evaluation task.

\paragraph{Limitations and future work.}
Our dataset comprises 2,619 samples from a single generation pipeline; extending coverage to diverse generators and artistic styles would improve generalization. The current 3-point rating scale, while reliable for inter-annotator agreement, limits the granularity of evaluation. Future work may explore finer-grained rubrics, native 3D representations as VLM input, and larger open-source VLMs to further narrow the gap with human judgment.

\section*{Ethical Use of Data and Informed Consent}
We follow standard research ethics throughout dataset creation and evaluation. All human participation was compensated fairly and based on informed consent. We do not collect sensitive personal information. We store data securely and restrict access to authorized researchers. We release the dataset under a license that prohibits misuse and requires appropriate attribution.

\section*{Acknowledgements}
We gratefully acknowledge our colleagues at Roblox, Kiran Bhat, Tinghui Zhou, Alvin Chan, and Michael Spedden, for their support of this work. We thank Kiran Bhat and Tinghui Zhou for their thoughtful paper reviews and valuable suggestions for improvement, and Alvin Chan and Michael Spedden for their contributions to the evaluation guidelines and the annotation exercise.
{
    \small
    \bibliographystyle{ieeenat_fullname}
    \bibliography{main}
}


\end{document}